\documentclass[runningheads]{llncs}

\usepackage[T1]{fontenc}
\def\doi#1{\href{https://doi.org/\detokenize{#1}}{\url{https://doi.org/\detokenize{#1}}}}
\usepackage{graphicx}
\usepackage[titlenumbered,ruled]{algorithm2e}
\usepackage{listings}
\usepackage{color}
\usepackage{multirow}

\newcommand{\revision}{\color{black}}

\usepackage{pifont}
\newcommand{\cmark}{\ding{51}}%
\newcommand{\xmark}{\ding{55}}%

\begin{document}
\title{Data-Driven Deep Supervision for Skin Lesion Classification}
%
%
\author{Suraj Mishra \inst{1} \and Yizhe Zhang \inst{2} \and Li Zhang \inst{3} \and Tianyu Zhang \inst{4} \and X. Sharon Hu \inst{1} \and  Danny Z. Chen\inst{1}}
\authorrunning{S. Mishra et al.}
%
\institute{University of Notre Dame, Notre Dame, USA \email{\{smishra3,shu,dchen\}@nd.edu} \and Nanjing University of Science and Technology, Nanjing, China \and Qingdao Women and Children's Hospital of Qingdao University, Qingdao, China \and University of Connecticut, Storrs, USA}
\maketitle              
\begin{abstract}
Automatic classification of pigmented, non-pigmented, and depigmented non-melanocytic skin lesions have garnered lots of attention in recent years. However, imaging variations in skin texture, lesion shape, depigmentation contrast, lighting condition, etc.~hinder robust feature extraction, affecting classification accuracy. In this paper, we propose a new deep neural network that exploits input data for robust feature extraction. Specifically, we analyze the convoutional network's behavior (field-of-view) to find the location of deep supervision for improved feature extraction. To achieve this, first we perform activation mapping to generate an object mask, highlighting the input regions most critical for classification output generation. Then the network layer whose layer-wise effective receptive field matches the approximated object shape in the object mask is selected as our focus for deep supervision. Utilizing different \textit{types} of convolutional feature extractors and classifiers on three melanoma detection datasets and two vitiligo detection datasets, we verify the effectiveness of our new method.   


\end{abstract}
%
%
%

\section{Introduction}
Automatic classification of skin lesions such as pigmented/non-pigmented lesions concerning skin cancers~\cite{isic16,isic17,rank2,rank5,rank7,rank10} and depigmented and hypomigmented lesions concerning vitiligo (vitiligo vs vitiligo-type, e.g., pityriasis alba, nevus depigmentous~\cite{viti,vit-liu}) in RGB images has garnered lots of research attention in recent years. Convolutional neural network (CNN) based models have shown promising results in accurate classification of skin lesions. However, imaging variations in skin texture, lesion shape, depigmentation contrast, lighting condition, etc.~hinder robust feature extraction, thus affecting classification accuracy. 

For input data specific robust feature extraction, ensuring the discriminativeness of the learned features in the hidden layers of a CNN is critical~\cite{mishra-vessel}. In
\cite{googlenet}, deep supervision was explored as an effective way to encourage feature maps at different layers to be directly predictive of the final output. By providing a companion objective using auxiliary classifiers~\cite{dsn}, deep supervision generates feedback for a local output (at a hidden layer). The combined error (i.e., global and local) is back-propagated, influencing the hidden layer update process and 
ensuring highly discriminative feature maps. But, where and how to apply deep supervision still remains an active research topic. Recently, Mishra~et al.~\cite{ds-tmi} studied data-driven deep supervision for medical image segmentation utilizing target object labels (or masks) as guidance for deep supervision. Yet, for image classification problems in general, the key issue of deep supervision location selection in a CNN becomes challenging due to the absence of object-level labels, since in classification problems, only image-level labels are normally available.

Previous work \cite{mishra-vessel,ds-tmi} exploited the observation that if the receptive field (RF) or field-of-view of a segmentation CNN matches in size the target object size in the input images, then the segmentation accuracy can be improved. Relative to the CNN's RF,
smaller objects are lost in the network’s sub-sampling operations while robust global features of larger objects are not well captured due to a smaller RF~\cite{ds-tmi}. We extend the observation/technique of deep supervision to tackle various skin lesion classification tasks, proposing data-driven deep supervision. 
Specifically, for deep supervision location selection, 
we utilize the input domain information (target object sizes) to identify the layer with preeminent contribution to feature extraction, and select this layer as the target location for deep supervision as auxiliary classification. Hence, as in~\cite{ds-tmi}, we employ an RF based approach to determine the preeminent layer in a classification CNN. But, note that for image classification problems, object-level labels are usually not available (only image-level labels, e.g., \{yes, no\}, are given). Thus, a difficulty to our approach is how to devise an effective method to estimate the sizes of objects in input images that possibly influence the classification results.

Each convolutional (conv) layer of a CNN extracts local features from a region of an input image to generate its 
output. The 
RF of the CNN is defined as the input region that directly affects the CNN's output~\cite{lecun}. By analyzing the near-\textit{Gaussian} distribution of RF, Luo et al.~\cite{erf} identified the unresponsive boundary regions of RF as non-output-affecting. The actual output-affecting region of the input image, which is smaller than the theoretical RF, is defined as the effective receptive field (ERF) of the network~\cite{erf}. Intuitively, convolutional filters extract improved features when the target object size (called morphological size) matches the filters’ ERF and the target lies at the center of the ERF~\cite{ds-tmi}.

From the viewpoint of ERF matching, for skin lesion classification using a deep CNN, we need to address two challenges: (1) determining the characteristic size of objects in the input images that the CNN 
is applied to; (2) determining the CNN’s ERF, whose size should match the characteristic object size. For the first challenge, we explore activation mapping to determine the morphological size of the input objects (approximate regions of interest in input images which have significant contribution to output generation). Since such \textit{regions of interest} can vary largely in size, for the second challenge, instead of using the fixed (or default) network ERF of a deep CNN, we utilize the layer-wise ERF (LERF)~\cite{mishra-vessel}. LERF represents the ERF of each conv layer in the CNN with respect to the image input. The conv layer with an LERF similar to the dataset-specific approximate object size is used as the target layer for deep supervision.

In data-driven deep supervision for skin lesion classification, we first approximate the morphological input object size from image-level labels using activation mapping. Then we introduce deep supervision from the layer of the network whose LERF size matches (as well as possible) the morphological object size. We validate our approach on three pigmented/non-pigmented lesion classification datasets concerning skin cancer (ISIC 2016~\cite{isic16}, ISIC 2017~\cite{isic17}, and ISIC 2018~{\revision \cite{isic18}}) and two depigmented/hypopigmented lesion classification datasets concerning vitiligo (vitiligo vs vitiligo-type, e.g., pityriasis alba, nevus depigmentous). For the public ISIC 2016, ISIC 2017, and ISIC 2018 datasets, we achieve comparable scores with the challenge winners. For both the vitiligo datasets~\cite{viti}, 
we outperform existing methods. 

\begin{figure}[t]
    \centering
    \includegraphics[width=0.78\textwidth]{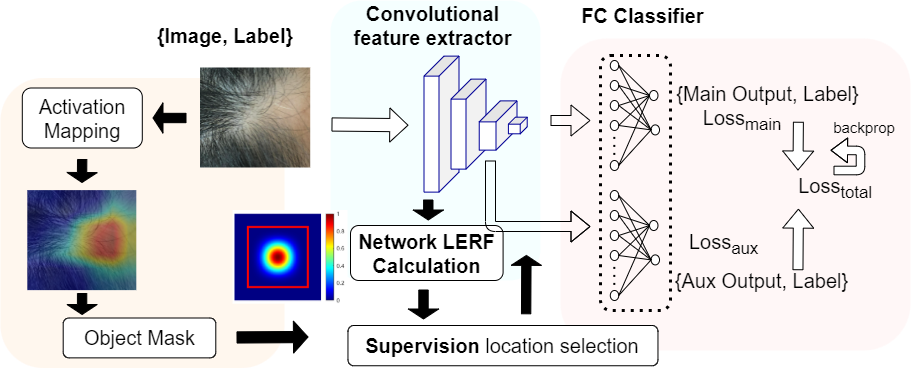}
    \caption{Our proposed framework for data-driven deep supervision for skin lesion classification. The major components of our framework are (1) LERF determination, (2) morphological object size approximation, and (3) deep supervision employment, shown in blue, orange, and red background, respectively. Specifically, utilizing activation maps, a morphological object mask is generated. The network layer whose LERF size well matches the morphological object size is chosen for deep supervision.}
    \label{fig:overall}
\end{figure}

\section{Method}
\label{sec:method}
The overall framework of our proposed data-driven deep supervision for skin lesion classification is shown in Fig.~\ref{fig:overall}. The major components of our framework are (1) LERF determination, (2) morphological object size approximation, and (3) deep-supervision employment. Section~\ref{ssec:lerf} and Section~\ref{ssec:obj} highlight LERF determination and morphological object size approximation, respectively. We then show how to use the information thus obtained for data-driven deep supervision in Section~\ref{ssec:dds}, along with some details of deep supervision implementation.    

\subsection{Layer-wise Effective Receptive Field (LERF)}
\label{ssec:lerf}
\begin{figure}[t]
    \centering
    \includegraphics[width=0.915\textwidth]{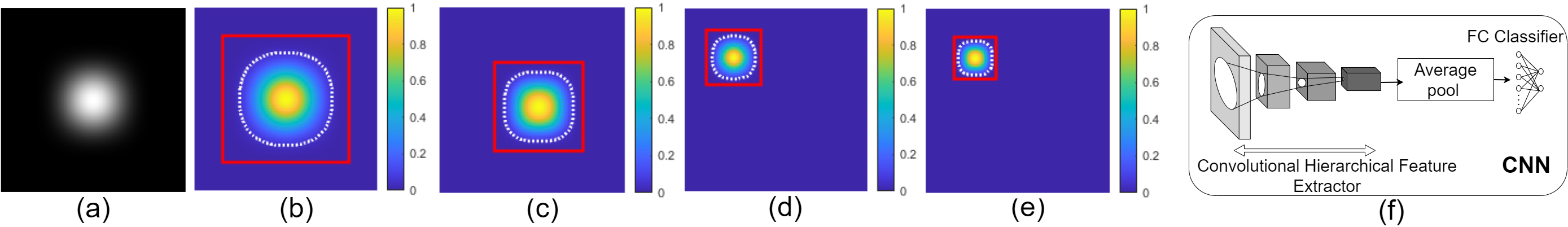}
    \caption{(a) Gaussian distribution of receptive field generated using the VGG13 architecture~\cite{vgg} with linear activations. (b) The boundary of RF is shown in red square. The boundary of ERF is shown in dashed white lines. Plots are generated for the last layer of the feature extractor (L34). (c)-(e) LERF plots are for L28, L24, and L21. (f) The extracted convolutional hierarchical features are average-pooled and classified.}
    \label{fig:erf}
\end{figure}

We take a partial derivative based approach to measure the influence on an input image region by an output node (a 
pixel) of a conv layer as in~\cite{erf,ds-tmi}. For a classification CNN with $N$ conv layers, $LERF_h$ is associated to layer $h$ ($h = 1,2,\ldots,N$). Our objective is to determine $\frac{\partial q_{i_t,j_t}}{\partial p_{i,j}}$, where $p_{i,j}$ is an input node (pixel) and $q_{i_t,j_t}$ is the target node at the output of the conv layer $h$ for which the influence measurement is being performed. Following~\cite{ds-tmi}, the region of influence (i.e., $\frac{\partial q_{i_t,j_t}}{\partial p_{i,j}}$) is determined using backpropagation (as all non-zero gradients on the input image). In Fig.~\ref{fig:erf}, LERFs associated to different layers of a representative CNN network are shown. Any node $q_{i_t,j_t}$ of a target layer can be selected for backpropagation since different nodes only generate translational shifts in the output (see Fig.~\ref{fig:erf}).

To extract the $LERF$ size from $\frac{\partial q_{i_t,j_t}}{\partial p_{i,j}}$, first we threshold (1 if $\frac{\partial q_{i_t,j_t}}{\partial p_{i,j}}$ > $th$; else 0) to remove the unresponsive boundary regions. Then, the pixel count of the active region is square-rooted to approximate the size of the $LERF$. Such a size approximation is inspired by the near-Gaussian distribution of $ERF$~\cite{ds-tmi}. 
Following \cite{erf}, a $th = 0.0455$ (i.e., a range of two standard deviations of the \textit{Gaussian} distribution, 4.55\% of the maximum value, i.e., 1 at the center) is used in $LERF$ calculation. In Fig.~\ref{fig:erf}, the convex hull of the $LERF$ pixels (the smallest convex area covering all the pixels) is marked with a closed white dotted curve. With non-linear activations (i.e., ReLU), this convex hull has a shape of a deformed square or an evolved ellipse. In the absence of network non-linearity, $LERF$ has a circular shape (see Fig.~\ref{fig:erf}). Hence, a square root based operation is used to approximate such a deformed shape of $LERF$ with network non-linearity.

\subsection{Object Size Approximation Using Activation Mapping}
\label{ssec:obj}
We explore an activation mapping based morphological object size approximation. Consider a classification CNN (generating an output label for an input image), which contains a stack of conv, activation, and sub-sampling layers for feature extraction. CNN-extracted features are average-pooled and fed to a fully connected (FC) classifier for final output generation (see Fig.~\ref{fig:erf}(f)).

\begin{figure}[t]
    \centering
    \includegraphics[width=0.93\textwidth]{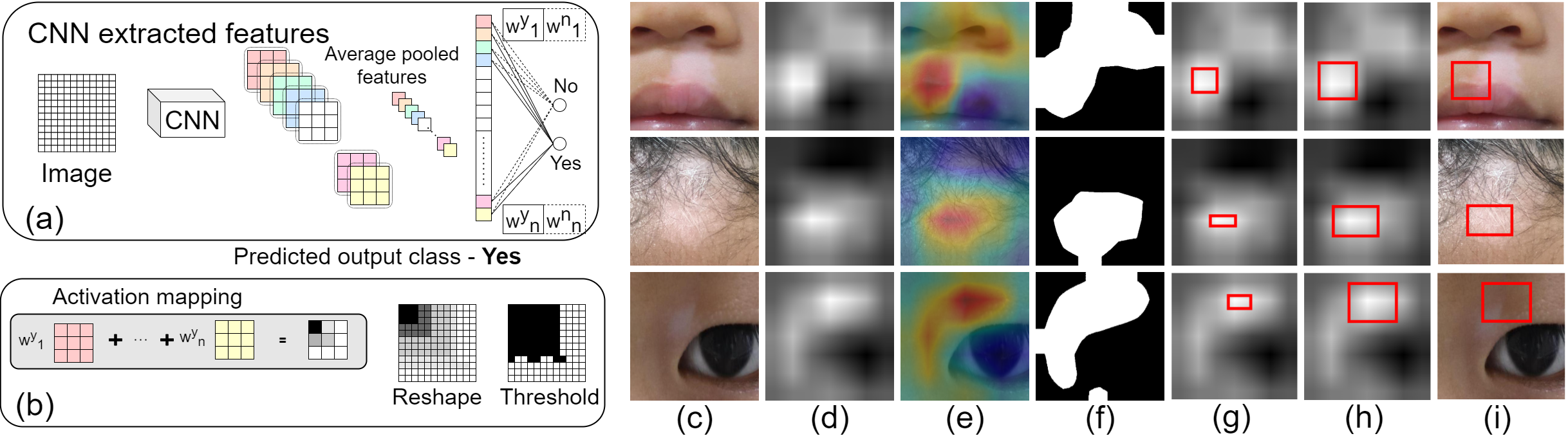}
    \caption{(a) A CNN with extracted conv features being pooled and classified for final output generation. (b) An activation map generated for a predicted class \textit{yes}. The generated activation map is thresholded  with $\delta = 0.5$ to generate a summary-mask. (c)-(d)-(e)-(f) Example images, activation maps in gray, activation maps in color, and summary-masks, respectively. (g) Bounding boxes for the central regions. (h)-(i) Object masks shown on activation maps and input images, respectively.}
    \label{fig:activation}
\end{figure}

For an input image, propagating through the CNN, let $f_k(i,j)$ represent the activation of a unit/channel/kernel $k$ in the last conv layer at spatial location ($i,j$). Then performing average pooling for unit $k$ results in $F_k = \sum_{i,j} f_k(i,j)$. Ignoring the bias term, the FC classifier layer with weights $w^c_k$ associated to a class $c$ generates output $\sum_k w^c_k F_k$ for class $c$. This output is used for loss calculation after a softmax-\textit{type} operation. By plugging the value of $F_k$ into the class score, we obtain $\sum_k w^c_k \sum_{i,j} f_k(i,j) = \sum_k \sum_{i,j} w^c_k f_k(i,j)$. In~\cite{cam}, the term $\sum_k w^c_k f_k(i,j)$ is defined as the activation map ($A_c$) for class $c$ (see Fig.~\ref{fig:activation}(b)), indicating the importance of the activation leading to the classification of an image for class $c$. As activation mapping indicates the importance of activations on an input image, we utilize it for object mask generation.

For object mask generation, we utilize the activation map $A_c$ and reshape it to the input image size using cubic-interpolation to obtain $A'_c$. The reshaped map $A'_c$ is then thresholded with a threshold $\delta'$ (i.e., $A'_c(i,j) = 0$ if $A'_c(i,j) < \delta'$; else $A'_c(i,j) = 1$). We use $\delta' = 0.9$ to determine the \textit{central region}. The smallest bounding box containing the central region is computed (see Fig.~\ref{fig:activation}(g)). Then we perform a region-growing type process to increase the size of the bounding box (equally in four directions) till we reach a pixel with a 50\% probability value ($\delta = 0.5$) in the activation map. A box in Fig.~\ref{fig:activation}(g) is increased in size till it touches the boundary of the summary mask (see Fig.~\ref{fig:activation}(f)), and the result is shown in Fig.~\ref{fig:activation}(h). Such processes are based on the special and stable properties of skin images (similar appearance, captured in similar conditions, using specific devices, etc). A common target lesion in skin images is a single connected region lying around the center of an image. With our approach, the most critical regions associated with the classification of the images in Fig.~\ref{fig:activation}(c) are shown in Fig.~\ref{fig:activation}(i). In Fig.~\ref{fig:activation}(h)-(i), example object masks are shown for some vitiligo images. It is interesting to note that the transition regions have the highest impact on final output generation. As in Sec.~\ref{ssec:lerf}, square root of the pixel count of an object mask (red bounding box regions in Fig.~\ref{fig:activation}(i)) is used as an approximate size of the object mask. The average size of all such object masks over sampled training set images is used as the morphological object size ($Obj$) of a specific dataset.

\subsection{Deep Supervision Employment}
\label{ssec:dds}

To apply deep supervision, we use $LERF_h$'s (computed in Section~\ref{ssec:lerf}) and $Obj$ (computed in Section~\ref{ssec:obj}) to find a target layer $L_{target}$, i.e., the conv layer with preeminent contribution to feature extraction. $L_{target}$ is taken as the layer $h^*$ with $|Obj\!-\!LERF_{h^*}| \leq |Obj\!-\!LERF_h|, \forall ~h \in\{1,\ldots,N\}$. Intuitively, $L_{target}$ is the conv layer in an $N$-layer CNN whose $LERF$ size best matches the $Obj$ size. To perform deep supervision using the output of $L_{target}$, we need to classify the features associated with $L_{target}$. The classification scores are then used for the auxiliary loss calculation. Our total loss is computed as $Loss_{total} = Loss_{main} + Loss_{aux}$, where $Loss_{main}$ is the main loss and $Loss_{aux}$ is the auxiliary loss.

\noindent \textbf{Remarks on Implementation.}
Here, we discuss two aspects related to deep supervision implementation: (1) auxiliary classifier design; (2) transfer learning. 

For deep supervision employment, conv features are processed by a classifier. The classified output is then used to compute the auxiliary loss. In this work, we explore two types of classifier architectures for auxiliary classifier design. (i) In a \textit{VGG}-type classifier~\cite{vgg}, multiple FC layers are used along with activation and dropout layers. In contrast, (ii) a \textit{ResNet}-type classifier~\cite{resnet} is light weight and uses a single FC layer as the classifier (details in Supplemental Material).     

Transfer learning plays a critical role (better/quicker optimization) in image classification by initiating the training process with a previously trained model~\cite{tl}, 
and has shown to be effective with improved classification accuracy. In data-driven deep supervision, an auxiliary loss is introduced from the preeminent layer of the network. Such model modifications encourage us to adopt a super-model based encapsulation approach, in which the base model is initialized by transfer learning and the auxiliary branch is initialized randomly (He-initialization~\cite{he_init}). Benefits of transfer learning are shown in Section \ref{sec:exp}.

\section{Experiments and Results}
\label{sec:exp}
\noindent
{\bf Datasets.}
Our experiments use the following datasets. \textbf{ISIC 2016}~\cite{isic16}: It contains 900 training and 379 test images. We randomly divide the original training set into 649 images for training and 251 images for validation. \textbf{ISIC 2017}~\cite{isic17}: It contains 2000 training, 150 validation, and 600 test images. Experiments are performed for task-3A, i.e., melanoma detection. \textbf{ISIC 2018}~\cite{isic18}: It contains 10,015 training, 193 validation, and 1512 test images with external evaluation. Considering the smaller size of the validation set, we randomly select 999 images ($\approx 10\%$) from the training set and add them to the validation set (9016 train and 1192 validation images). \textbf{Vitiligo (in-house)}: It contains 2188 images (1227 training, 308 validation, and 653 test images). {\revision The in-house dataset consists of images from retrospective consecutive outpatients obtained by the dermatology department of Qingdao Women and Children’s Hospital (QWCH) in China. For each patient with suspected vitiligo (e.g., pityriasis alba, hypopigmented nevus), clinical photographs of the affected skin areas were taken by medical assistants using a point-and-shoot camera (as described in~\cite{viti}).} \textbf{Vitiligo (public)}~\cite{viti}: It contains 1341 images (672 training, 268 validation, and 401 test images).

\noindent
{\bf Experimental Setup.}
We resize all the images of each dataset to 224$\times$224. To reduce overfitting on a small training set, data augmentation is performed using random flipping and cropping. In cropping, a random portion of an image is extracted and resized to the target size (224$\times$224). We use a standard back-propagation implementing \textit{Adam}~\cite{adam} with a fixed learning rate of 0.00002 ($\beta_1 = 0.9, \beta_2 = 0.999$, and $\epsilon = 1\mathrm{e}{-8}$). Experiments are performed on NVIDIA-TITAN and Tesla P100 GPUs, using PyTorch for $1k$ epochs. The batch size for each case is selected as the maximum size allowed by the GPU. We use a network initialization as discussed in Section \ref{ssec:dds}. For all the datasets, cross-entropy loss is used as $Loss_{main}$ and $Loss_{aux}$. The best performing model version on the validation set is saved as the model checkpoint.

VGG13~\cite{vgg} and ResNet18~\cite{resnet} are used as the backbone models. Untrained networks with \textit{He}-initialization~\cite{he_init} are used for the $LERF$ experiments. To neutralize variations in $LERF$ computation caused by network non-linearity, the mean over 20 iterations is used. We should note that the LERF computing time is much less than the CNN training time (less than a minute on an NVIDIA P100 machine). Following the processes presented in Section~\ref{ssec:obj}, $Obj$ for resized images is determined. Using the $LERF$ values of the network layers, deep supervision locations are decided. ResNet-type classifiers are used as the default architecture for the auxiliary classifier (more 
details 
in Supplemental Material).

\begin{table*}[t]
\caption{Lesion classification results on skin image datasets. {\revision Proposed*/Proposed --- without/with deep supervision.}}
\begin{center}
\scalebox{0.82}{
\begin{tabular}{| c | c  c  c  c |  c | c  c  c  c |}
\hline
\multicolumn{5}{|c|}{ISIC 2016 Dataset} &
\multicolumn{5}{c|}{ISIC 2017 Dataset}\\
\hline
\rule{0pt}{1pt} Method & AUC & Acc  & Sen & Spe & Method & AUC & Acc  & Sen & Spe \\ \hline
VGG-16~\cite{vgg} &  0.826 & 0.826 & 0.413 & 0.928 & Galdran~\textit{et al.}~\cite{rank10} & 0.765 & 0.480 & \textbf{0.906} & 0.377 \\ 
GoogleNet~\cite{googlenet} & 0.801 & 0.847 & 0.507 & 0.931 & Vasconcelos~\textit{et al.}~\cite{rank7} & 0.791 & 0.830 & 0.171 & 0.990 \\
DRN-50~\cite{drn50} &  0.783 & 0.855 & 0.547 & 0.931 & Yang~\textit{et al.}~\cite{rank5} & 0.830 & 0.830 & 0.436 & 0.925\\
Gutman~\textit{et al.}~\cite{isic16} & 0.804 & 0.855 & 0.507 & 0.941 & Diaz~\textit{et al.}~\cite{rank2} & 0.856 & 0.823 & 0.103 & \textbf{0.998} \\
ARDT-DenseNet~\cite{ardt} & 0.837 & 0.857 & \textbf{0.816} & 0.756 & ARDT-DenseNet~\cite{ardt} & \textbf{0.879} & \textbf{0.868} & 0.668 & 0.896 \\
SDL~\cite{sdl2019} & 0.829 & \textbf{0.857} & \textbf{-} & - & SDL~\cite{sdl2019} & 0.830 & 0.830 & - & - \\
\hline
Proposed* - VGG{\revision 13} & 0.830 & 0.852 & 0.334 & 0.980 & Proposed* - VGG{\revision 13} & 0.808 & 0.845 & 0.325 & 0.971 \\
Proposed - VGG{\revision 13} & 0.837 & 0.871 & 0.387 & \textbf{0.990} & Proposed - VGG{\revision 13} & 0.831 & 0.860 & 0.590 & 0.926 \\
\hline
Proposed* - RES18  & 0.817 & 0.850 & 0.400 & 0.961 & Proposed* - RES18 & 0.786 & 0.825 & 0.282 & 0.956\\
Proposed - RES18 & \textbf{0.850} & 0.842 & 0.440 & 0.951 & Proposed - RES18 & 0.791 & 0.848 & 0.470 & 0.940 \\
\hline
\hline
\multicolumn{5}{|c|}{Vitiligo (in-house) Dataset} &
\multicolumn{5}{c|}{Vitiligo (public) Dataset}\\
\hline
\rule{0pt}{1pt} Method & AUC & Acc  & Sen & Spe & Method & AUC & Acc  & Sen & Spe \\ \hline
VGG-13~\cite{vgg} & 0.851 & - & 0.791 & 0.913 & VGG-13~\cite{vgg} & 0.995 & - & 0.972 & 0.963\\
ResNet~\cite{resnet} &  0.840 & - & 0.775 & 0.902 & ResNet~\cite{resnet} & 0.958 & - & 0.952 & 0.957\\
DenseNet~\cite{densenet} & 0.847 & - & 0.784 & 0.906 & DenseNet~\cite{densenet} & 0.982 & - & 0.962 & 0.961\\
Dermatologists~\cite{viti} & - & - & \textbf{0.811} & \textbf{0.999} & Dermatologists~\cite{viti} & - & - & 0.964 & 0.803 \\ 
\hline
Proposed* - VGG13  & 0.852 & 0.882 & 0.797 & 0.907 & Proposed* - VGG13 & 0.970 & 0.973 & 0.992 & 0.951 \\
Proposed - VGG13 & \textbf{0.931} & \textbf{0.904} & 0.777 & 0.941 & Proposed - VGG13 & \textbf{0.998} & \textbf{0.988} & \textbf{0.996} & \textbf{0.975} \\
\hline
Proposed* - RES18  & 0.897 & 0.865 & 0.662 & 0.925 & Proposed* - RES18 & 0.987 & 0.950 & 0.962 & 0.932 \\
Proposed - RES18 & 0.920 & 0.893 & 0.757 & 0.933 & Proposed - RES18 & 0.990 & 0.968 & 0.962 & 0.975\\
\hline
\end{tabular}}
\end{center}
\label{tab:results}
\end{table*}

\noindent
{\bf Results and Discussions.}
Table~\ref{tab:results} shows the results.
For each dataset, the \textit{Proposed*} or \textit{Proposed} row gives our results obtained without or with deep supervision. The backbone CNN networks used in each experiments (i.e., VGG13 and RES18) are shown.
Applying data-driven deep supervision indeed improves classification accuracy. On the ISIC 2016 dataset, our method exceeds the results of the challenege winner~\cite{isic16}. On ISIC 2017, our results are comparable to the best performing models in the leader-board with published results~\cite{rank2,rank5,rank7,rank10}. 

\begin{figure}[t]
    \centering
    \includegraphics[width=0.93\textwidth]{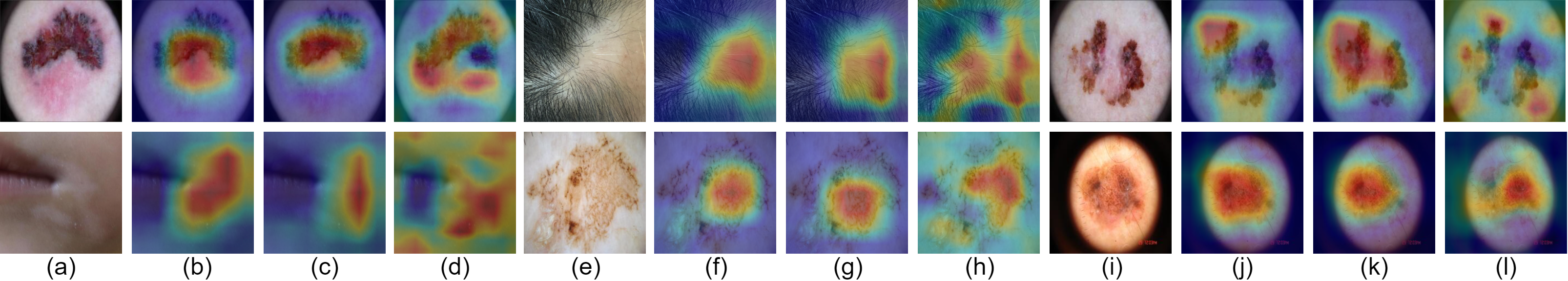}
    \caption{(a)-(e)-(i) Example images. (b)-(f)-(j) Activation maps generated without deep supervision. (c)-(g)-(k) Activation maps by deep-supervision (main branch). (d)-(h)-(l) Activation maps by deep-supervision (aux branch).}
    \label{fig:activation_change}
\end{figure}

On the \textbf{ISIC 2018} dataset, external evaluation is performed and we receive the results as an ``overall score'' (aggregated over all the 7 classes), i.e., no Acc, Sen, Spe, and AUC results for it. Our \textit{Proposed*}-VGG13 achieves an overall score of 0.664 (comparable to the challenge rank 50~\cite{isic18rank50}); our \textit{Proposed}-VGG13 yields an overall score of 0.701 (comparable to the challenge rank 10~\cite{isic18rank10} among the published results without using external data, with the best reported score being 0.845~\cite{isic18rank1}). For our \textit{Proposed*}-RES18, an overall score of 0.626 is received. With supervision, our \textit{Proposed}-RES18 achieves an overall score of 0.661. These highlight the benefits of our deep supervision. 
To further examine the impact of data-driven deep supervision on skin lesion classification, we compare changes in activation maps with deep supervision. In Fig.~\ref{fig:activation_change}, changes in activations due to deep supervision for example images are shown. 

\begin{table*}[t]
\caption{Ablation study analysis. {\revision C: Classifier type (R({\it ResNet-type})/V({\it VGG-type})); DS: Deep supervision location; TL: Transfer learning (with(\cmark)/without(\xmark))}.}
\begin{center}
\normalsize
\scalebox{0.7}{
\begin{tabular}{| c  c  c  c | c  c  c  c | c  c  c  c | c c c c |}
\hline
\multicolumn{4}{|c|}{Method} &
\multicolumn{4}{c|}{ISIC 2016 Dataset} &
\multicolumn{4}{c|}{Vitiligo (in-house) Dataset} &
\multicolumn{4}{c|}{Vitiligo (public) Dataset}
\\
\hline
\rule{0pt}{1pt} Backbone & C & DS & TL & AUC & Acc  & Sen & Spe & AUC & Acc  & Sen & Spe & AUC & Acc & Sen & Spe\\ \hline
VGG13 & R & \xmark & \cmark & 0.830 & 0.852 & 0.334 & 0.980  & 0.852 & 0.882 & 0.797 & 0.907 & 0.970 & 0.973 & 0.992 & 0.951\\
VGG13 & R & L28 & \cmark & 0.837 & \textbf{0.871} & 0.387 & \textbf{0.990} & 0.931 & 0.904 & 0.777 & 0.941 & \textbf{0.998} & \textbf{0.988} & \textbf{0.996} & \textbf{0.975}\\
\hline
VGG13 & R & L24 & \cmark & 0.830 & 0.839 & 0.293 & 0.974 & 0.924 & \textbf{0.907} & \textbf{0.791} & 0.941 & - & - & - & -\\
 VGG13 & R & L24\&L28 & \cmark & 0.834 & 0.847 & 0.320 & 0.977 & 0.931 & 0.897 & 0.716 & \textbf{0.950} & - & - & - & -\\
 VGG13 & R & L31 & \cmark & 0.838 & 0.852 & 0.387 & 0.967 & \textbf{0.936} & 0.891 & 0.811 & 0.915 & 0.997 & 0.973 & 0.983 & 0.957\\
\hline
VGG13 & R & \xmark & \xmark  & 0.715 & 0.810 & 0.227 & 0.954 & 0.809 & 0.860 & 0.703 & 0.915 & 0.892 & 0.905 & 0.962 & 0.821\\
VGG13 & R & L28 & \xmark & 0.744 & 0.829 & 0.213 & 0.980 & 0.842 & 0.862 & 0.804 & 0.879 & 0.921 & 0.928 & 0.954 & 0.889\\
\hline
VGG13 & V & L28 & \cmark & \textbf{0.840} & 0.847 &\textbf{0.400} & 0.957 & 0.917 & 0.891 & 0.703 & 0.947 & \textbf{0.998} & 0.980 & 0.992 & 0.963\\
\hline
\end{tabular}}
\end{center}
\label{tab:ablation}
\end{table*}

\noindent
{\bf Ablation Study.}
Using ISIC 2016, Vitiligo (in-house), and Vitiligo (public) as example cases, we conduct an ablation study to examine the contributions of deep supervision location selection, classifier architecture, and transfer learning. Table~\ref{tab:ablation} shows the results. \textbf{Deep Supervision Location Selection:} Contribution of our deep supervision is shown by the DS column in Table~\ref{tab:ablation}.
In addition to data-driven deep supervision (at L28 in VGG13), experiments are conducted by applying deep supervision at different layers (L24, L24 \& L28, and L31). On the Vitiligo (in-house) dataset, better AUC and specificity are observed for deep supervision with the L31, and L24 \& L28 case, respectively.
\textbf{Transfer Learning:}
Contribution of transfer learning is shown in the TL column of Table~\ref{tab:ablation}. In the absence of transfer learning, there is a significant drop in all accuracy metrics. \textbf{Classifier Architecture:}
Experiments using VGG-type classifier architectures are discussed in Section \ref{ssec:dds}. Both the classifier architectures generate comparative results (in column C of Table~\ref{tab:ablation}).

\section{Conclusions}
In this paper, we proposed a new data-driven deep supervision approach for skin lesion classification. Utilizing CNNs' layer-wise ERF information and input object size approximated by activation mapping, deep supervision location is selected. 
Experiments on various datasets verify the effectiveness of our approach.

{\revision In future work, we plan to improve the robustness of our proposed method by substituting CAM~\cite{cam} with improved frameworks such as grad-CAM~\cite{gradcam} and score-CAM~\cite{scorecam}. Further, the single target lesion assumption is a limitation of the proposed method. We plan to extend our method to multiple target objects of different sizes based on the data-driven deep supervision segmentation method for multiple object sizes as in~\cite{ds-tmi,Decomp-Integrate-2019,mishra_isbi21}.}

\vspace{0.1in}
{\revision \noindent\textbf{Acknowledgement.} 

\noindent
This work was supported in part by NSF grant CCF-1617735.}

%
%
%
\bibliographystyle{splncs04}
\bibliography{mybib}
\end{document}


\title{Supplemental Material for Submission \#1816\\
``Data-Driven Deep Supervision for 
Skin Lesion Classification''}
%
%
%
\authorrunning{- et al.}
%
%
\maketitle              

\begin{figure}[ht]
    \centering
    \includegraphics[width=1.15\textwidth]{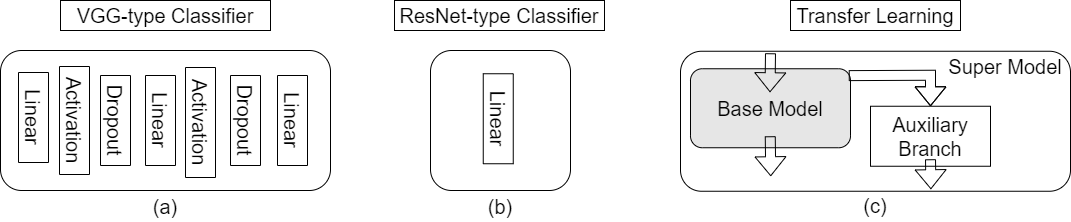}
    \caption{(a) A VGG-type classifier with multiple fully-connected layers. (b) A ResNet-type classifier with a single fully-connected layer. (c) Our super model based encapsulation approach for transfer learning implementation for the modified network architecture with deep supervision.}
    \label{fig:remarks}
\end{figure}

\begin{figure}[ht]
    \centering
    \includegraphics[width=0.80\textwidth]{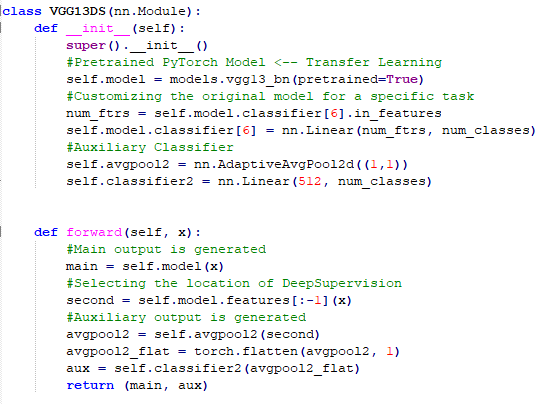}
    \caption{PyTorch implementation of our encapsulated model with auxiliary branch.}
    \label{fig:encaps}
\end{figure}

\begin{table*}[ht]
\caption{Approximated object size (Obj) for different datasets.}
\begin{center}
\normalsize
\scalebox{0.9}{
\begin{tabular}{ c | c | c | c | c | c }
\hline
\multicolumn{1}{c|}{-} &
\multicolumn{1}{c|}{ISIC 2016} &
\multicolumn{1}{c|}{ISIC 2017} &
\multicolumn{1}{c|}{ISIC 2018} &
\multicolumn{1}{c|}{Vitiligo (in-house)} &
\multicolumn{1}{c}{Vitiligo (public)}
\\
\hline
\rule{0pt}{1pt} Obj & 67.19 & 71.53 & 66.07 & 67.86 & 66.31 \\ \hline
\end{tabular}}
\end{center}
\label{tab:ablation}
\vspace{-0.1in}
\end{table*}

\begin{figure}[ht]
    \centering
    \includegraphics[width=0.85\textwidth]{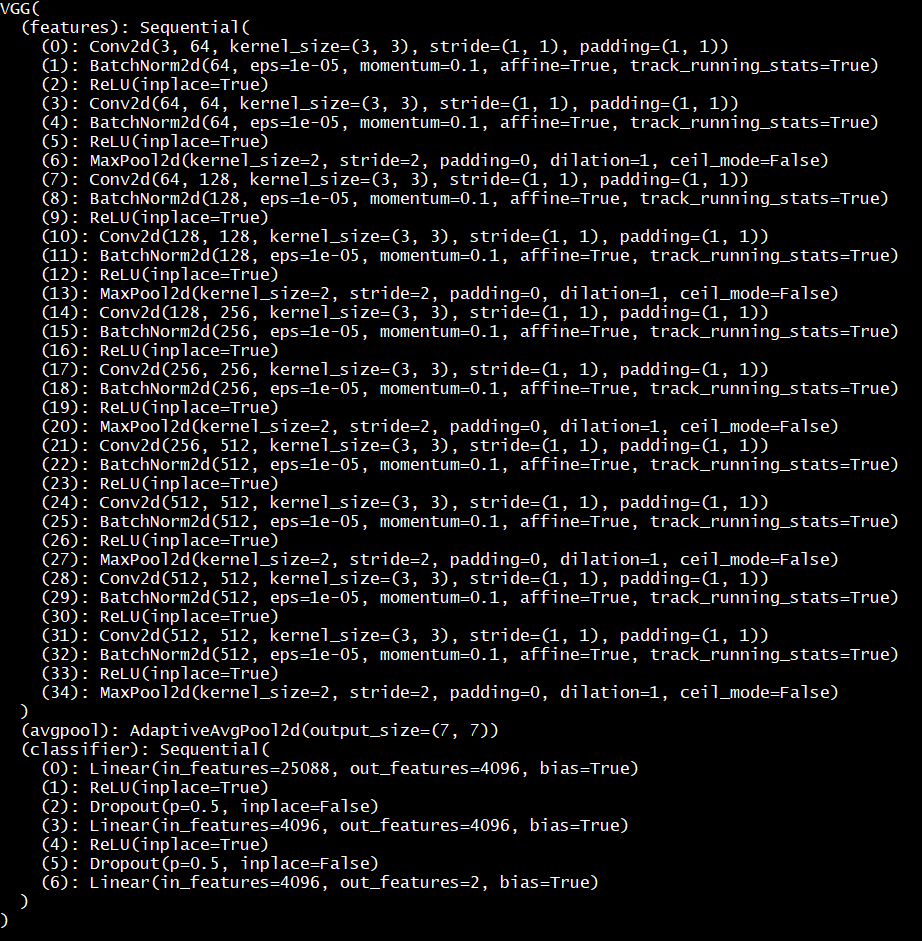}
    \caption{VGG13 layer arrangement.}
    \label{fig:vgg_arch}
\end{figure}

%